# Contextually Customized Video Summaries via Natural Language

Jinsoo Choi
KAIST EE
jinsc37@kaist.ac.kr

Tae-Hyun Oh
MIT CSAIL
thoh.mit.edu@gmail.com

In So Kweon
KAIST EE
iskweon@kaist.ac.kr

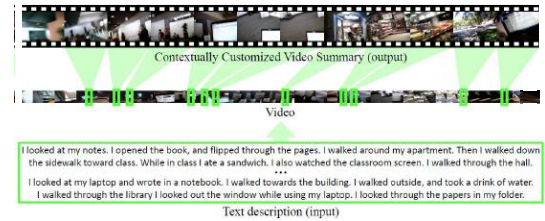

Figure 1: Our algorithm generates contextually customized video summaries through simple user text descriptions. Given a preferred summary description, our method extracts and temporally aligns the semantically relevant segments in a video according to its context.

## Abstract

*The best summary of a long video differs among different people due to its highly subjective nature. Even for the same person, the best summary may change with time or mood. In this paper, we introduce the task of generating contextually customized video summaries through simple text. First, we train a deep architecture to effectively learn semantic embeddings of video frames by leveraging the abundance of image-caption data via a progressive manner, whereby our algorithm is able to select semantically relevant video segments for a contextually meaningful video summary, given a user-specific text description or even a single sentence. In order to evaluate our customized video summaries, we conduct experimental comparison with baseline methods that utilize ground-truth information. Despite the challenging baselines, our method still manages to show comparable or even exceeding performance. We also demonstrate that our method is able to automatically generate semantically diverse video summaries even without any text input.*

## 1. Introduction

It is a great irony that as more memory storage becomes available, the importance of summarizing becomes more prevalent. Today, we are able to collect hours worth of videos without worrying about the lack of memory, which has enabled people to store more memories and experiences. Also, we are exposed to visual media being uploaded on the web nonstop. These phenomena has led to the desire to automatically extract only meaningful parts from this ever-growing media [4, 28]. As an effect, video summarization has gathered much attention from not only academia, but also from corporations. In the computer vision community, there have been numerous approaches to summarizing videos, many of which define what aspects are important in a video to automatically extract such keyframes or subshots. Some recent works attempt to learn what aspects are important from example video summaries.

Video summarization is a highly subjective task [11, 42] which cannot be left up to a single algorithm's decision. In this paper, we do not attempt to define nor learn what aspects are important in a video, but instead learn the context of videos through text descriptions paired with visual data, allowing the user to define what is desired as a summary in natural language. Our method can generate customized video summaries reflecting the semantics expressed via text which will differ dramatically among different people and even change for the same person with time or emotions. When the user decides to provide minimal text or even no text at all, our method is still able to produce semantically coherent and diverse video summaries.

As an overview of our approach, we first jointly learn semantic embeddings of video frames and its sentence descriptions. We initially learn semantic representations from an abundant image-caption data. Then, we progressively transfer this knowledge across to the video-caption domain by learning the residual domain knowledge between the image-caption and video-caption domains. As a result, we are able to effectively learn semantically rich video frame and sentence representations despite the relatively smaller video-caption dataset. Based on the learned embeddings, our method extracts the semantically relevant video frames given the user-specific summary description. The actual video summary is then produced by combining the video segments containing the relevant frames and preserving the temporal order by utilizing the hidden Markov model (HMM) and a decoding method based on the Forward-backward algorithm. We evaluate our method components in comparison to challenging baselines both quantitatively and qualitatively.

Our main contributions are as follows: (1) We introduce a contextually customized video summarization method via natural language. Our task differs from that of previous video summarization methods in that it involves summarizing hours worth of videos with user-specific texts. (2) We learn a pairwise ranking model with a progressive and residual training method to learn valuable information from abundant image-caption data and effectively learn from relatively smaller and domain specific video-caption data which proves to learn more rich representations compared to fine-tuning results. By virtue of our emphasis on learning rich semantic representations, our method is also able to produce semantically coherent summaries with little or no text input. (3) We utilize the HMM and a marginal posterior decoding method based on the Forward-backward algorithm to produce quality summaries which are temporally aligned. (4) To evaluate the subjective aspect of video summarization with user-specific texts, we develop comparison baselines which directly uses embeddings of the ground truth text annotations. In such challenging comparisons, our method still manages to produce comparable or exceeding performances.

## 2. Related Work

Video summarization methods have been characterized by how to define summarization criteria for *better* or *representative* summaries. Early works have mainly developed model-based approaches which exploit low-level visual features such as motion [37], background stitching [1] and spatio-temporal features [16] to find interesting keyframes or subshots. Based on an interestingness metric, model-based approaches compose summaries with an emphasis on simplicity or diversity of the subshots depending on the model representations such as graph [25] and compact coreset representation [28].

Summaries generated by low-level features or model constraints rarely retain high level context information. More recent works focus on retaining semantic information for better human correspondence: important objects (Lee *et al*. [17]), object tracks (Liu *et al*. [20]), motion based summaries via semantic object context (Oh *et al*. [26]), and object relationships (Lu *et al*. [21]). While these methods deal with higher level context information, they only consider a single crafted criterion to produce summaries.

In order to make up for the shortfall of hand crafted summarization criteria, recent works introduce various data driven approaches. These approaches build a computational model that learns human summary preferences from data. Supervised methods use summary annotations obtained from human workers in the form of ground truth summaries [7, 9], highlight annotations [41] or GIF-formatted summaries [10]. On the other hand, Khosla *et al*. [13] and Yang *et al*. [40] leverage unsupervised learning techniques to mine representativeness from web data. However, these models learn from reference summaries, leading to a general algorithm that may neglect subjective preferences.

As mentioned, video summarization is a highly subjective task. We categorize video summarization approaches reflecting subjectiveness into implicit or explicit methods, *i.e.* data driven or user interaction-based. For a data driven approach, Sharghi *et al*. [31] propose a noun-based video summarization where the summaries are learned via reference summaries specific to a predefined set of noun classes. Users have control over which nouns (up to 3 nouns) to include in the summary, but not over specific summary composition which is learned via reference summaries. Also, this approach is not applicable to novel nouns other than the predefined set of nouns. A data driven approach [27] concurrent to our work takes a reference text summary as input to generate a video summary. It assumes temporally aligned texts, and trains an objective function comprising of predefined terms. While both methods are conceptually similar, our method focuses on the semantic aspect of summarization. Song *et al*. [33] propose a title based video summarization. Their idea is to query video titles through web search engines to crawl similar concept images, and analyze them jointly with the video to indirectly reveal canonical concepts with hand-crafted visual features. Our method deals with multi-modality in an end-to-end manner.

As user interaction-based approaches, Goldman *et al*. [6] propose to render action summary layouts via a series of user annotations, and Han *et al*. [11] propose to propagate subshots out of user specified keyframes. These methods require the user to traverse through the video frames in order to reflect personal summary preferences. By contrast, our approach allows a user to describe one's personally preferred summary via text. Thus, the user does not need to iteratively interact with video frames, leading to high efficiency in terms of time and human effort for customized video summarization. Our contextually customized video summarization approach is motivated by recent developments on joint embeddings of images and captions [3, 5, 22], while the goal is different in that our multi-modal embedding module aims to learn contextual domain knowledge from a relatively smaller and domain specific video-caption data. Textual descriptions itself is fully semantic and contextual, which our algorithm aims to naturally reflect in our video summaries.

## 3. Customizing Video Summaries via Text

We introduce a method which takes a simple text description as input and generates a video summary reflecting its context. Firstly, our goal is to find a set of frames semantically relevant to the input text. In order to do this, we learn a semantic embedding function that jointly maps frames and sentences to a common embedding space. Once the set of semantically relevant frames are selected, the video summary is generated by combining the subshots containing the selected frames. By utilizing the hidden Markov model, temporal alignment is induced on the summary.

**Progressive-residual embedding** Given a text description, we aim to extract semantically relevant frames from a whole video. Common practice would be to train or fine-tune a deep model with a video-caption dataset. However, we instead leverage an image-caption dataset (source domain) to transfer useful knowledge not found in video-caption datasets (target domain) in a progressive and residual manner. As opposed to recent video-caption datasets, large image-caption datasets such as MS COCO [19] contain multiple detailed sentence descriptions per image. This enables training with diverse textual descriptions, leading to not only a model learning rich semantical relations between image and language, but also a model robust to subjective textual descriptions. Since a video frame is essentially an image, we can leverage insightful information from the image-caption domain.

Our deep semantic embedding model is a pairwise ranking model [30, 39] (*i.e.* triplet network) which learns the similarity between frame (image) and text modalities. Our model is trained on triplet data $t^{(i)}=(y^{(i)},x_+^{(i)},x_-^{(i)})$ where $y^{(i)}$ is the anchor sentence, and $(x_+^{(i)},x_-^{(i)})$ are the positive and negative frames respectively. The hinge loss for a triplet $(y^{(i)},x_+^{(i)},x_-^{(i)})$ is defined as:

$$l(y^{(i)}, x_+^{(i)}, x_-^{(i)}) = \max\{0, m - s(f(x_+^{(i)}), g(y^{(i)})) + s(f(x_-^{(i)}), g(y^{(i)}))\}, \quad (1)$$

where $m$ is a margin parameter that regularizes the margin between positive and negative pairs. The functions $f(\cdot)$ and $g(\cdot)$ denote embedding functions for frames and sentences respectively. The compatibility scoring function $s(\cdot,\cdot)$ is the cosine similarity scoring function.

Our model is progressively trained, which the image and sentence features trained with an image-caption dataset are transferred into our model. An overview of our progressively trained model is shown in Fig. 2. Basically, we train a two-column progressive network where the first column is a deep multimodal architecture such as the $m$-CNN [22] or order embedding network [38] trained in the image-caption domain. The second column consists of fully connected layers which take the frame and sentence features as input (video-caption domain). Here, the feature outputs from the first column is transferred via lateral connections so that the second column is able to access rich knowledge already learned from the first column.

What is different from our progressive model to the orig-inal progressive neural network [29] is that our model only transfers the output feature from the first column rather than all of its layer activations. Also, the transferred *features* are *frozen* which cannot be transformed by the second column. These differences in fact give rise to residual task learning, meaning that the second column is explicitly let to fit a residual mapping across columns. The residual task learning we mention here is slightly different from that of the deep residual network of He *et al*. [12] in that our task involves learning the (scaled) residual mapping across columns, whereas deep residual networks involve residual learning adopted to every few stacked layers. In this way, the first column most likely learns most of the rich semantic representations and the second column learns the *residual* knowledge such as frame-caption semantics specific to the target domain. Formally, the embedding features learned by our method can be defined as:

$$v_t = \mathcal{F}(h_t, \{W\}) + \alpha v_s, \quad (2)$$

where $v_t$ and $v_s$ denote the embedding features from the second and first columns (*i.e.* target and source domain columns) respectively. The function $\mathcal{F}(h_t, \{W\})$ represents the residual mapping to be learned with frame/sentence feature input $h_t$ and weights $W$. The value $\alpha$ is a learned scalar, initialized by a random small value.

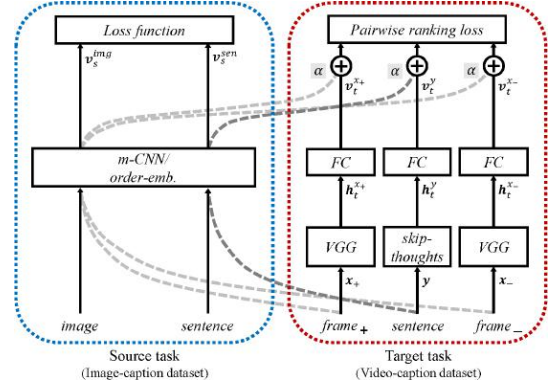

Figure 2: Overview of our progressively trained model.

Since the second column is explicitly designed to learn the residual information, a relatively smaller network such as the fully connected layer is sufficient enough. This is shown in the experiments where our method generally performs better than the baselines. Also, since the second column consists of fully connected layers, it does not introduce any noticeable additional parameters nor computational complexity during training compared to simple fine-tuning if not less. Another advantage of our progressive model is that since most of the semantic representation is learned from the first column which is transferred to the second column, we are able to learn even with a relatively small video-caption data and still learn rich representations.

**Video summary construction** An intuitive way to construct video summaries based on a text description is to assign a relevant subshot to each sentence in the text description, while also preserving the temporal order of subshot events. Thus, our goal is to select the best possible matching subshot for each sentence so that the subsequent subshots are temporally aligned. In other words, we would like to select just enough most relevant frames for each sentence, and only consider those frames to construct a temporally aligned sequence of subshots containing the selected frames. We can achieve all this by the hidden Markov model (HMM). We simply model the summary transition behavior in the following HMM structure.

– The observation space $\mathcal{Y}=\{y_1, y_2, \cdots, y_N\}$ is the set of text description sentences.

– The state space $\mathcal{Z}=\{z_1, z_2, \cdots, z_F\}$ is all the sampled frames of an input video.

– Sequence of observations $\boldsymbol{O}=[o_1, o_2, \cdots, o_T]$ is the description sentence sequence in order, where $T=N$ and $o_i=i$, thus $\boldsymbol{O}=[1, \cdots, N]$.

– Sequence of states $\mathbf{Q}=[q_1, \cdots, q_T]$ is a hidden variable which is regarded as generating the observations, where $q_t \in \mathcal{Z}$. This is what we ultimately want to find from given observations above.

– State transition probability matrix $\boldsymbol{A} \in \mathbb{R}^{F-1 \times F}$ of which entries $\boldsymbol{A}_{ij}$ stores the transition probability from the state $z_i$ to the state $z_j$, $p(q_t=z_j|q_{t-1}=z_i)$. We set $\boldsymbol{A}_{ij}=\frac{1}{F-i}$ for $j>i$, and otherwise 0. This transition matrix has non-zero entries for $\boldsymbol{A}_{i<j}$, row-wise normalized. By this matrix, transition can only occur forward in time, enforcing the result sequence of frames to be temporally aligned, but without any detailed transition preference.

– Observation probability (or emission) matrix $\boldsymbol{B} = \left[ \vec{B}_1, \vec{B}_2 \cdots \vec{B}_N \right] \in F \times N$ is a matrix where $\boldsymbol{B}_{ij}$ stores the probability of observing $y_j$ from the state $z_i$, $p(o_t=y_j|q_t=z_i)$. Here, only the top-$k$ compatibility scores among $F$ frames between a sentence is kept at the corresponding positions while other entries are assigned as zeros. Thus, the column vectors $\vec{B}_j$ of the emission matrix have non-zero score entries only at top-$k$ scoring frame positions for each observation $o_j$. In other words, for each given sentence, $k$ determines the number of candidate frames to consider for selection. If for example, the top-$k$ frames to be considered at $o_{j+1}$ does not contain frames that come temporally after top-$k$ frames at $o_j$, temporal transitions will not occur due to the forward transition constraint of $\boldsymbol{A}$. Thus, the value $k$ is incremented iteratively until all forward transitions take place, and thus is the minimal number of top scoring frames per sentence required to produce a temporally aligned sequence prediction. The emission matrix is row-wise normalized.

|  | Ours$_{m\text{-CNN}}$ | | Ours$_{\text{order}}$ | |
|---|---|---|---|---|
|  | mAP | mAD | mAP | mAD |
| **Marginal by FB** | **25.00** | **1.81** | **24.17** | **1.62** |
| MAP by Viterbi | 20.83 | 5.49 | 24.17 | 2.39 |
| DTW | 11.67 | 8.33 | 11.67 | 9.95 |

Table 1: Summary performance comparison according to decoding on egocentric dataset.

Given a specified HMM model, learning the model parameters is not necessary, but rather simply decoding the probable hidden state sequence is our goal. The most common method for decoding the most likely sequence of states from the given observation is through the Viterbi algorithm. The Viterbi algorithm finds the maximum a posteriori (MAP) state sequence such that

$$\mathbf{Q}^* = \arg\max_{\mathbf{Q}} p(\mathbf{Q}|\mathbf{O}) = \arg\max_{\mathbf{Q}} p(\mathbf{Q},\mathbf{O}). \quad (3)$$

However, as shown in Table 1, we empirically found that finding the sequence of most probable states from the marginal posterior gives better results than the Viterbi-based MAP solution for our task. We pose the problem as

$$q_t^* = \arg\max_{q_t} p(q_t|\mathbf{O}) = \arg\max_{q_t} p(q_t,\mathbf{O}). \quad (4)$$

Here, the marginal probability $p(q_t,\mathbf{O})$ is computed by marginalizing the joint distribution $p(\mathbf{Q},\mathbf{O})$ over all possible permutations of $\{q_{i \in \{1,\cdots,T\}/t}\}$. However, the brute-force marginalization will take $O(F^N N)$ complexity which is intractable. In order to efficiently compute $p(q_t,\mathbf{O})$, we use a simple trick to decompose $p(q_t,\mathbf{O})$. Let us denote $\mathbf{O}_{i:j} = [o_i, o_{i+1}, \cdots, o_j]$, where $j>i$. Then,

$$p(q_t, \mathbf{O}) = p(\mathbf{O}_{1:t}, \mathbf{O}_{t+1:T}|q_t)p(q_t)$$
$$= p(\mathbf{O}_{1:t}|q_t)p(\mathbf{O}_{t+1:T}|q_t)p(q_t) = p(\mathbf{O}_{1:t}, q_t)p(\mathbf{O}_{t+1:T}|q_t).$$

where the second equality is by independence of $o_i$'s.

This decomposition can be seen as just another representation, but interestingly the terms $p(\mathbf{O}_{1:t}, q_t)$ and $p(\mathbf{O}_{t+1:T}|q_t)$ can be efficiently computed by the conventional Forward and Backward (FB) algorithms [24] respectively. This only takes $O(F^2 N)$ for computing $p(q_t, \mathbf{O})$, and further reduces to $O(kFN)$ in our case due to the $k$-sparsity of the emission matrix $\boldsymbol{B}$. Now we efficiently obtain the most likely states $q_t$ for $t \in \{1, \cdots, N\}$ by Eq. (4), which corresponds to the resulting video summary.

A probable rational behind the better performance of this marginal posterior method is that the marginalization in the Markov process has a smoothing effect over neighboring information [23, 24], while the Viterbi algorithm greedily maxes out a specific value during decoding that could be sensitive to noise [24].

## 4. Experiments

We analyze our method and its components using two datasets: the Egocentric daily life dataset [17] and the TV episodes dataset [42]. The egocentric dataset consists of 4 extremely diverse videos captured from head-mounted cameras, lasting for 3-5 hours each — a total of over 17 hours of video. The videos contain scenarios such as eating, working, driving, shopping, cooking, *etc*. in various locations and view points. The TV episodes dataset consists of 4 television episodes of 40 minutes each. These videos contain even more diverse scenarios and dramatic situations which shift frequently within the relatively short 40-minute duration. All training and values presented are results of leave-one-out cross-validation for each video in the datasets.

We divide our experiment into 3 parts in order to evaluate our method component qualities.
- *Evaluating embedding quality*: We evaluate the embedding quality by considering the task as a retrieval problem, measuring the recall-at-K and median rank.
- *Evaluating summary quality*: To evaluate overall video summary quality, we measure the mean Average Precision (mAP) and a distance metric from the ground truth, in comparison to some challenging baselines.
- *Generating summary with less text*: We show that our method can generate semantically coherent and diverse video summaries via single sentence or no text inputs.

Additional qualitative results can be found in the supplementary material.

### 4.1. Evaluating embedding quality

To assess the embedding quality of video frames and sentences, we regard quality assessment as a retrieval task, as done in [22, 38]. Unlike image-caption datasets which involve multiple (usually 5) elaborate sentences directly describing what is seen in the image, the aforementioned datasets involve single sentence descriptions for a collection of frames in simple terms. Sentences in the egocentric dataset describe video segments in terms of what the first-person is doing, instead of what is seen. For example, a video segment showing the inside of a room is labeled with a single sentence "I looked around the room," rather than describing what is actually seen in the room. For the case of TV episodes dataset, the majority of video segments involves characters speaking to each other about a specific topic, which is vaguely captioned. For instance, a scene is described as "Rigsby recognizes Hanson, and discusses his past marriage" which involves specific prior knowledge and more importantly, does not sufficiently describe the visual scene. In this sense, retrieval using these datasets is a highly challenging task.

The retrieval results on egocentric and TV episodes datasets are shown in Table 2. We adopt a standard metric for retrieval assessment, recall-at-K (R@K) where K values are selected depending on the scale of test frames and sentences, shown at the top-left entry of each subtable, and on which would sufficiently express retrieval quality. The scale of test frames and sentences shown are averaged over all videos in the dataset. The median rank (Med. r) is reported in top-percentile manner in order to take the test data scale into account.

**Baseline methods** We compare our model with a number of state-of-the-art models with varied training or fine-tuning schemes: (1) a pairwise ranking model (triplet network) with VGG-19 [32] image features and skip-thought [15] sentence vectors as inputs, (2) an $m$-CNN model [22] trained from scratch, (3) $m$-CNN model fine-tuned (whole model), (4) $m$-CNN model fine-tuned (last output layer only), (5) an order-embedding model [38] trained from scratch, (6) order-embedding fine-tuned (whole model), (7) order-embedding fine-tuned (last output layer only). Our progressive-residual embeddings are trained by having either $m$-CNN or order-embedding models as the first column, and a pairwise ranking model as the second column with VGG-19 and skip-thought feature inputs. Our embeddings are learned by training the first column with an image-caption dataset MS COCO [19], and training the second column with the egocentric or TV episodes datasets where frames are sampled every second. The fine-tuned baselines are pre-trained with the MS COCO and fine-tuned with the egocentric or TV episodes datasets as well.

**Training details** All of the embeddings are set to 1024-dimensional vectors. We used the margin $m=0.2$ in Eq. (1), 50 contrastive terms for each positive data pair and trained for 10 epochs while saving the model whenever an improvement occurred on the development set. We implemented with Theano [2] and used the ADAM [14] for optimization and mini-batch size of 100. Fig. 3 illustrates the embeddings from baselines and our method.

Our learned embeddings (*i.e.* ProgRes$_{mCNN/order}$) show better performances for the majority of R@K metrics compared to other fine-tuned baselines, and consistently show better results for median rank. We also provide performances of the baselines trained with C3D [35] clip features which directly trains with video clip segment and sentence pairs (shown in Table 3). Notice that the networks trained with frame/sentence pairs show better performances, especially for text-to-visual retrieval, which is actually the main concern for our method.

### 4.2. Evaluating summary quality

From each dataset, 3 reference summary texts (written by different individuals) for each of the 4 videos are given. These texts are each composed of 24 sentences to which 5-second video subshots are each assigned, resulting in 2-minute ground truth video summaries. In consensus to the dataset, we also uniformly sample frames every 5 seconds and apply our proposed summary construction algorithm[1]. Evaluation is simply done by measuring how similar the predicted video summaries are with the ground truth summaries. We use the popular performance metric, mean Average Precision (mAP) [34]. In addition to this popular metric we propose another metric that reveals another aspect of video summary quality. This metric measures the average of the normalized temporal distance between ground truth reference summary segments and its closest segment from the predicted summary. We call this metric as the mean Average Distance (mAD). While mAP measures the degree of overlap between the predicted summary and reference, mAD reflects how temporally close the predicted summary segments are to the reference summary segments. This allows for a more semantic evaluation since temporally close segments most likely have similar semantics. Note that reference summary texts are different from ground truth text annotations, where annotations are used for training, while testing is done solely on reference texts.

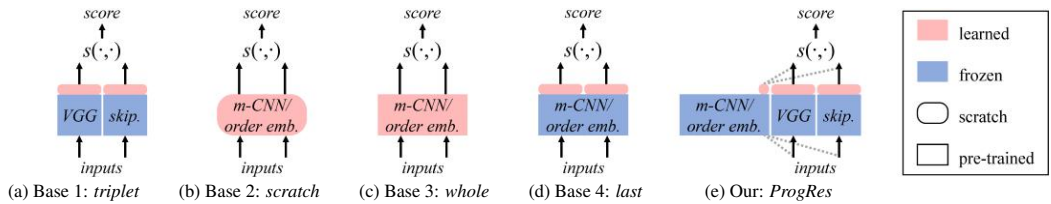

Figure 3: Embeddings from baselines and our model. Embeddings from (a) a pairwise ranking model (triplet network) trained on the target task with VGG-19 skip-thought features; (b) an embedding model (either $m$-CNN or order-embedding) trained from scratch; (c) fine-tuned model (whole); (d) fine-tuned model (last layer only); and (e) our model.

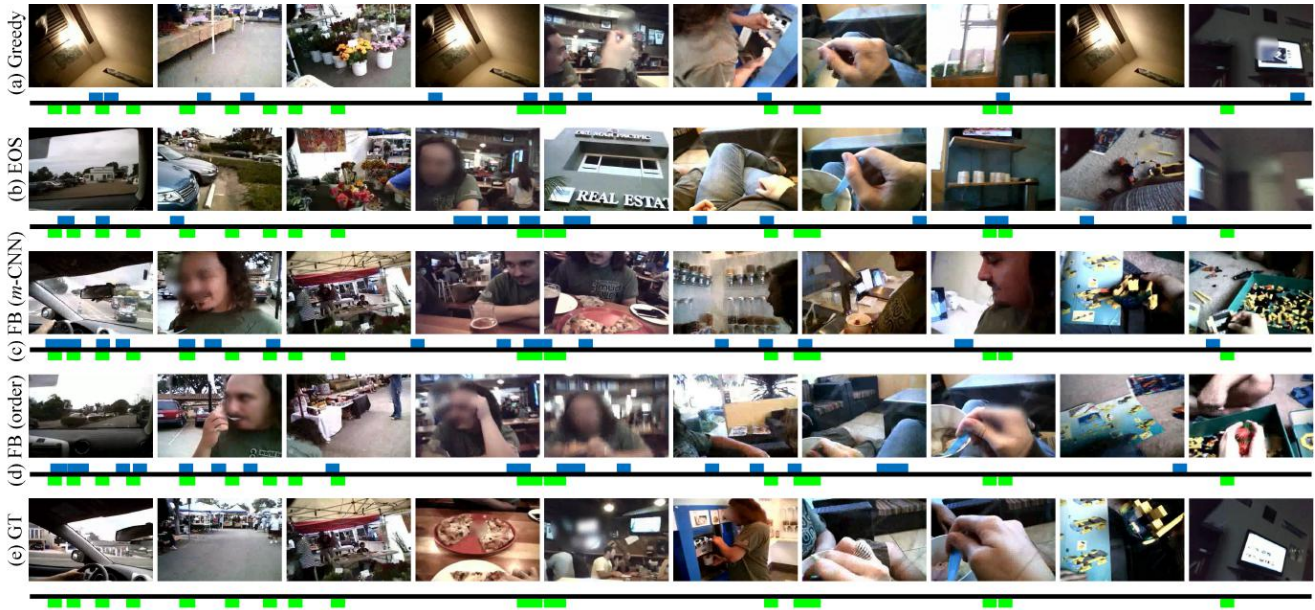

Figure 4: Summaries from baselines and our method. The blue and green blocks indicate relative subshot positions of the predicted summaries and ground truth reference summary respectively.

| #sen: 1516, #frm: 15267 | | Egocentric dataset | | | | | | | | | |
|---|---|---|---|---|---|---|---|---|---|---|---|
| | | Frame-to-Text | | | | | Text-to-Frame | | | | |
| | | R@1 | R@8 | R@64 | R@512 | Med. r (%) | R@1 | R@10 | R@100 | R@1000 | Med. r (%) |
| Frame-based | VGG-skip. triplet | 0.50 | 4.46 | 22.91 | 74.67 | 15.88 | 0.59 | 3.69 | 17.49 | 57.84 | 6.57 |
| | $m$-CNN $_{scratch}$ | 0.26 | 3.58 | 18.51 | 65.46 | 26.53 | 0.47 | 3.42 | 15.50 | 47.87 | 10.05 |
| | $m$-CNN $_{whole}$ | 0.32 | 2.89 | 16.90 | 61.76 | 26.25 | 0.59 | 3.19 | 14.89 | 47.83 | 9.23 |
| | $m$-CNN $_{last}$ | 0.47 | 4.18 | 17.66 | 65.84 | 22.42 | 0.62 | 2.96 | 15.36 | 54.77 | 6.05 |
| | ProgRes$_{mCNN}$ | 0.67 | 4.79 | 22.76 | 75.36 | 16.62 | 0.75 | 4.26 | 19.08 | 59.61 | 6.05 |
| | order-emb. $_{scratch}$ | 0.12 | 1.04 | 13.65 | 61.59 | 26.14 | 0.59 | 3.59 | 16.92 | 52.12 | 7.49 |
| | order-emb. $_{whole}$ | 0.15 | 1.31 | 13.42 | 58.17 | 28.06 | 0.80 | 3.86 | 16.98 | 52.75 | 7.49 |
| | order-emb. $_{last}$ | 0.34 | 3.87 | 16.82 | 62.08 | 25.48 | 0.58 | 3.05 | 15.60 | 53.11 | 6.99 |
| | ProgRes$_{order}$ | 0.62 | 5.13 | 23.96 | 74.11 | 17.03 | 0.73 | 4.20 | 19.42 | 60.02 | 5.38 |
| #sen: 225, #frm: 2584 | | TV episodes dataset | | | | | | | | | |
| | | Frame-to-Text | | | | | Text-to-Frame | | | | |
| | | R@1 | R@8 | R@64 | R@512 | Med. r (%) | R@1 | R@10 | R@100 | R@1000 | Med. r (%) |
| Frame-based | VGG-skip. triplet | 0.94 | 5.22 | 37.03 | - | 41.67 | 0.64 | 4.53 | 27.67 | 90.74 | 10.38 |
| | $m$-CNN $_{scratch}$ | 0.61 | 4.36 | 34.97 | - | 43.53 | 0.76 | 4.49 | 22.95 | 84.91 | 13.34 |
| | $m$-CNN $_{whole}$ | 0.39 | 4.74 | 33.23 | - | 45.71 | 0.36 | 3.50 | 24.23 | 86.14 | 12.41 |
| | $m$-CNN $_{last}$ | 0.85 | 5.34 | 35.51 | - | 43.72 | 0.41 | 4.71 | 26.66 | 88.28 | 10.74 |
| | ProgRes$_{mCNN}$ | 1.22 | 5.90 | 38.58 | - | 40.24 | 0.85 | 5.97 | 30.08 | 91.50 | 8.74 |
| | order-emb. $_{scratch}$ | 0.51 | 5.17 | 36.71 | - | 42.31 | 0.85 | 5.84 | 27.32 | 87.88 | 10.91 |
| | order-emb. $_{whole}$ | 0.76 | 5.42 | 36.52 | - | 42.19 | 0.99 | 5.60 | 28.26 | 87.63 | 10.30 |
| | order-emb. $_{last}$ | 1.16 | 5.94 | 35.46 | - | 44.02 | 0.42 | 4.89 | 24.89 | 86.71 | 11.87 |
| | ProgRes$_{order}$ | 0.97 | 6.34 | 38.42 | - | 40.49 | 1.22 | 5.62 | 28.71 | 87.61 | 9.21 |

Table 2: Ranking results on egocentric and TV episodes datasets. The best and runner-up results are shown in red and blue respectively. For R@K, a higher value is better, while a lower value is better for Med. r (%).

| #sen: 1516, #clip: 3053 | | Egocentric dataset | |
| --- | --- | --- | --- |
| | | Clip-to-Txt Med. r (%) | Txt-to-Clip Med. r (%) |
| Clip-based | VGG-skip. triplet | 17.61 / **15.88** | 16.11 / **6.57** |
| | $m$-CNN $_{scratch}$ | 29.72 / **26.53** | 24.43 / **10.05** |
| | $m$-CNN $_{last}$ | 26.99 / **22.42** | 21.17 / **6.05** |
| | order-emb. $_{scratch}$ | **25.76** / 26.14 | 19.96 / **7.49** |
| | order-emb. $_{last}$ | 26.57 / **25.48** | 22.14 / **6.99** |
| #sen: 225, #clip: 258 | | TV episodes dataset | |
| | | Clip-to-Txt Med. r (%) | Txt-to-Clip Med. r (%) |
| Clip-based | VGG-skip. triplet | **38.87** / 41.67 | 39.89 / **10.38** |
| | $m$-CNN $_{scratch}$ | 48.57 / **43.53** | 46.90 / **13.34** |
| | $m$-CNN $_{last}$ | **41.36** / 43.72 | 41.10 / **10.74** |
| | order-emb. $_{scratch}$ | **39.32** / 42.31 | 40.21 / **10.91** |
| | order-emb. $_{last}$ | 46.42 / **44.02** | 43.09 / **11.87** |

Table 3: Median rank results on egocentric and TV episodes datasets with C3D features. The previously shown frame-based results are attached as reference (best in bold).

| | | Egocentric | | TV episodes | |
| --- | --- | --- | --- | --- | --- |
| | | mAP | mAD | mAP | mAD |
| Base | Greedy | 10.00 | 17.58 | <span style="color:red">43.75</span> | 7.79 |
| Base | EOS | 20.83 | 2.65 | 8.33 | 19.54 |
| Our | FB (ProgRes$_{mCNN}$) | <span style="color:red">25.00</span> | 1.81 | 10.42 | <span style="color:blue">3.33</span> |
| Our | FB (ProgRes$_{order}$) | <span style="color:blue">24.17</span> | <span style="color:red">1.62</span> | <span style="color:blue">16.67</span> | <span style="color:red">2.69</span> |

Table 4: Video summarization quality results on egocentric and TV episodes datasets. The best and runner-up results are shown in red and blue respectively.

Since this work addresses the task of generating a video summary customized by a reference text, we need to develop comparison baselines that also produce a customized video summary given a reference text for fair evaluation. The baselines we develop take a reference text as input and matches its constituent sentences directly to the ground truth sentence annotations of the video. This closely resembles the experimental setup done in [42]. In this way, the baselines combine the video segments corresponding to the matched text annotations to generate textually customized video summaries. The following is the detailed explanation of the baselines:

1. *Greedy Text-embedding Selection*: The ground truth sentences as well as the reference text are embedded to skip-thought vectors [15]. Subshots are greedily selected by matching the reference summary text embeddings with the ground truth.

2. *Embedding-based Ordered Subshot*: Subshots are selected by matching the reference summary text skip-thought vectors with the ground truth, but by preserving the temporal order via dynamic programming approach.

The baselines essentially work on the text modality, and thus the source of error may be from text embedding quality. Despite our efforts to devise baselines for fair comparisons, the baselines in fact have more advantage over our method since they directly use the ground truth text annotations and work with a single modality domain, *i.e.* text. However, in this challenging comparison, our method still manages to show comparable or exceeding results.

Our method aims for personally customized video summarization, and thus we exclude reference summaries from our evaluation that entirely reuse sentences from the ground truth sentence annotations. The summarization quality results are shown in Table 4. Our methods both based on ProgRes$_{mCNN}$ and ProgRes$_{order}$ embeddings manage to produce better results in terms of all metrics on the egocentric dataset. On the TV episodes dataset, the greedy baseline performs best followed by our methods in terms of mAP. On the other hand, our methods show better performance than the baselines in terms of mAD. We point out that the reference summary texts of the TV episodes dataset closely resembles the ground truth text annotations if not identical. This may have caused high confidence matches leading to high mAP for the baselines since they are text-based, whereas our method is based on semantics.

It is worth noting that our method can produce video summaries *semantically* relevant to the provided text descriptions. Because of this, our method performs better than the baselines in all aspects on the egocentric dataset which contains diverse semantical reference texts. Furthermore, our method consistently performs the best in terms of mAD on the TV episodes dataset despite its near identical reference texts to the ground truth text annotations. An overview of the predicted summaries for baselines and our method is shown in Fig. 4.

---
[1]The TV episodes dataset reference texts involves 12 sentences and 10-second subshots, to which we adjust our samples accordingly.

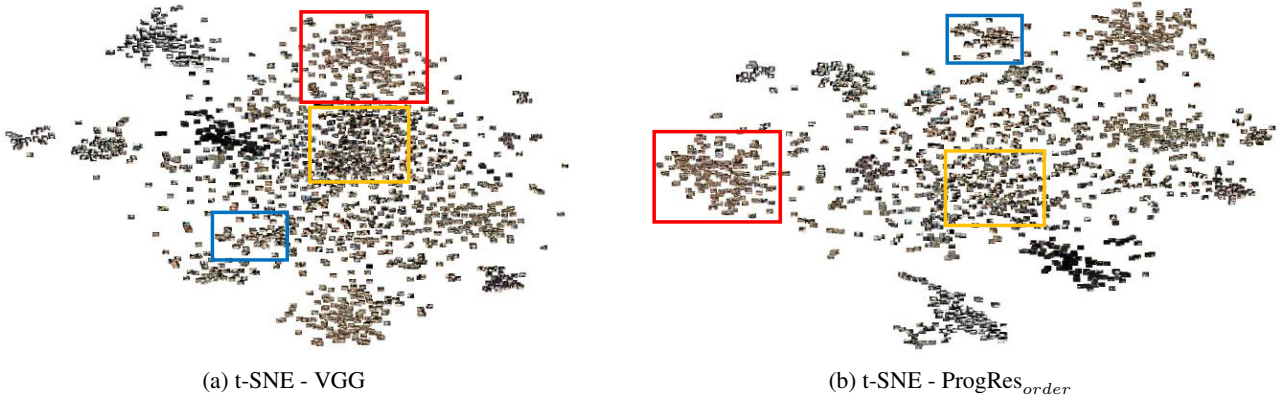

(a) t-SNE - VGG    (b) t-SNE - ProgRes$_{order}$

Figure 5: Barnes-Hut t-SNE visualization of (a) VGG-19 and (b) our ProgRes$_{order}$ embeddings. Best viewed when zoomed in.

|      |                       | mAP   | mAD   |      |                        | mAP   | mAD  |
|------|-----------------------|-------|-------|------|------------------------|-------|------|
| Base | Greedy                | 0.83  | 65.02 | Base | Uniform                | 18.75 | 0.99 |
| Base | EOS                   | 6.67  | 43.45 | Base | VMMR (VGG)             | 28.12 | 1.31 |
| Our  | FB (ProgRes$_{mCNN}$) | 12.50 | 10.17 | Our  | VMMR (ProgRes$_{mCNN}$)| 31.25 | 1.45 |
| Our  | FB (ProgRes$_{order}$)| 13.33 | 9.66  | Our  | VMMR (ProgRes$_{order}$)| 30.21 | 1.45 |
|      | (a) Single sentence   |       |       |      | (b) Without text       |       |      |

Table 5: Summary results using single sentence and without text.

### 4.3. Generating summaries with less text

We demonstrate how our learned video frame embeddings can produce semantically diverse summaries when given a single sentence or no text at all. We evaluate our algorithm on the egocentric dataset due to its rich context.[2]

**Summaries with single sentence input** We evaluate with a single sentence as input. The single sentence inputs are made based on the original multi-sentence reference summaries.[3] In turn, these single sentence inputs act as prior for contextually coherent video summarization. Summaries can be generated with single sentences without modifying our algorithm at all. The single sentence is simply duplicated multiple times (24 for the egocentric dataset) as input to our algorithm. Using the same ground-truths as the previous sections, our model consistently demonstrates better results on all metrics as shown in Table 5-(a).

**Summaries without text input** When a text input is not given at all, our model is able to generate semantically diverse summaries only via the visual embeddings learned. The video summary is generated by combining the subshots that contain the keyframes chosen by the Video-MMR [18] algorithm. Basically, a keyframe that is similar to the frames not yet selected while different from already selected keyframes is chosen at each iteration.

We compare Video-MMR applied to our ProgRes$_{mCNN}$ and ProgRes$_{order}$ visual embeddings to that of VGG-19 features. We also compare with the uniform sampling method which performs surprisingly well quantitatively. Previous works on video summarization such as [9] have shown that uniform sampling actually returns near state-of-the-art quantitative results, but show poor qualitative performance. In this sense, we present uniform sampling quantitative results for relative reference. The results are shown in Table 5-(b). Our method performs the best in terms of mAP while comparable in terms of mAD. For a visual quality assessment, we provide Barnes-Hut t-SNE visualization [36] of VGG-19 and our semantic embeddings of a test-split video frames in Fig. 5. Notice how the clusters in VGG-19 plot are relatively close together due to the visually oriented embeddings, whereas the supposedly same clusters in our ProgRes$_{order}$ plot are further away due to difference in semantics.

### 5. Conclusion

Language is an inherent ability of humans, making it one of the most natural means of communication and interaction. In this era of unlimited media and elaborate means, we present customized video summarization via language. We leverage the abundance of images and its captions to effectively learn joint embeddings of video frames and text. Our algorithm is able to generate temporally aligned video summaries instilled with the semantics written by individuals. For a task as subjective as video summarization, our work presents a means of customization with an ability inherent to humans.

**Acknowledgment.** This work was supported by Institute for Information & communications Technology Promotion(IITP) grant funded by the Korea government(MSIT) (2017-0-01780), and by the Technology Innovation Program (No. 10048320), funded by the Ministry of Trade, Industry & Energy (MI, Korea).

---

[2] The SumMe dataset [8] could be used as an evaluation dataset, but it has a relatively limited context change (mean duration: 2m40s) which may not be sufficient for evaluation in our scenario.

[3] Given the original multi-sentence reference summaries, we asked a subject to reduce redundant words from adjacent sentences and connect them with transition words to form single sentence inputs.